\documentclass[journal]{IEEEtran}
\ifCLASSINFOpdf
  \usepackage[pdftex]{graphicx}
  \usepackage{subfig}
\else
\fi
\begin{document}
%
\title{Extracting dispersion curves from ambient noise correlations using deep learning}
%
%
%

\author{\IEEEauthorblockN{Xiaotian Zhang, Zhe Jia, Zachary E. Ross, and Robert W. Clayton}
        \\\IEEEauthorblockA{California Institute of Technology}}

%
%

\markboth{IEEE Transactions on Geoscience and Remote Sensing}%
{Shell \MakeLowercase{\textit{et al.}}: Bare Demo of IEEEtran.cls for IEEE Journals}
%



\maketitle

\begin{abstract}
We present a machine-learning approach to classifying the phases of surface wave dispersion curves. Standard FTAN analysis of surfaces observed on an array of receivers is converted to an image, of which, each pixel is classified as fundamental mode, first overtone, or noise. We use a convolutional neural network (U-net) architecture with a supervised learning objective and incorporate transfer learning. The training is initially performed with synthetic data to learn coarse structure, followed by fine-tuning of the network using approximately 10\% of the real data based on human classification. The results show that the machine classification is nearly identical to the human picked phases. Expanding the method to process multiple images at once did not improve the performance. The developed technique will faciliate automated processing of large dispersion curve datasets.
\end{abstract}


%
\IEEEpeerreviewmaketitle

\section{Introduction}
The inversion of surface waves has become a standard method for determining the near-surface shear velocity. One reason for this is that surface waves can be relatively easily extracted from ambient noise correlations, and hence are not dependent on a suitable distribution of earthquakes. Another reason is that surface waves only require coverage over a 2D plane and not a 3D volume, which is what would be required for body waves (S-waves). This becomes important when dealing with dense seismic arrays that typically have a short deployment time.

The method usually consists of three steps, with the first being the determination of dispersion curves, which are measurements of the velocity as a function of frequency. Once this is done, a tomographic method is used to convert these line measurements into maps of the phase or group velocity as a function of frequency \cite{shapiro2005high}. The final step is to then convert velocity as a function of frequency at each (x,y)-point, to velocity as a function of depth, thus making a 3D model of the subsurface velocity. In this paper we will focus on applying machine learning to the first step in the process – determining the dispersion curves.

A commonly used procedure to extract the dispersion curves is the FTAN (Frequency-Time Analysis) method \cite{levshin1972frequency,herrmann2013computer}, in which the correlated signal between two stations is filtered by a sequence of zero-phase narrowband filters to determine the travel time of the surface waves as a function of frequency. Knowing the separation distance of the stations allows these measurements to be converted to phase velocity. If the envelope of the signal is used instead of the seismograms themselves then the group velocity is determined.

 At a given frequency, there may be a number of modes present that correspond to different eigenfunctions (dependencies with depth). The fundamental mode is the slowest mode with the over-tones increasing in velocity as the eigenfunctions penetrate deeper in depth. A key part of determining the dispersion curve is “picking” the travel time or equivalently the velocity since the distance is known.  This is similar to problem of picking P- and S-waves in determining earthquake locations, but here, the various surface-wave modes need to be classified for the inversion process. We typically pick the fundamental and 1st-overtone modes, and occasionally the 2nd-overtone if it can be seen. The more that are picked, the better the resolution of the resulting shear velocity model.
 
Picking the dispersion curves is very labor intensive, particularly when dealing with dense arrays. The motivation for automating this procedure is not only the large volume of data that is now available (an example of which is shown in Figure 1), but also the increased precision that is now required because of the density of stations. The process can be machine-assisted by defining target zones for the curves, but the output needs to be checked and adjusted because of spurious noise within these zones. Developing an automatic method to determine the dispersion curves is the subject of this paper. 

In recent years, deep learning has become state-of-the-art in numerous areas of artificial intelligence, which has quickly translated into major advances within seismology. Such applications include detection and picking of seismic waves \cite{ross2018p,zhu2019phasenet}, signal denoising \cite{zhu2019seismic}, and phase association \cite{ross2019phaselink}. These problems can all be cast as supervised learning objectives and benefit from the wealth of labeled datasets that exist in the seismological community. They bear structural similarities to that of dispersion curve picking, motivating the application of deep neural networks. In this study, we develop a deep learning approach to the dispersion picking problem with the goal of classifying tentative picks as fundamental mode, 1st overtone, or noise. Our approach uses deep convolutional networks to learn a low dimensional representation of the data which can be used for pixelwise segmentation of dispersion curves. We show that our approach can reliably and efficiently classify picks, which will greatly facilitate automated processing of large seismic datasets.

\begin{figure*} 
    \centering
  \subfloat{%
       \includegraphics[width=0.45\linewidth]{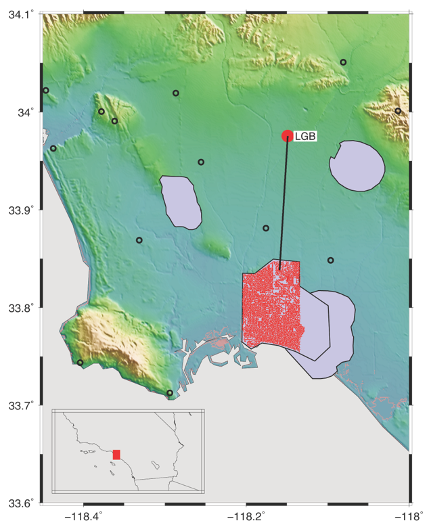}}
    \hfill
  \subfloat{%
        \includegraphics[width=0.55\linewidth]{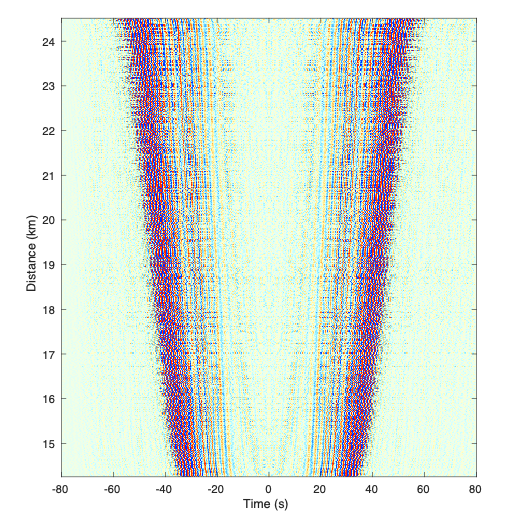}}
  \caption{Location and Correlations. The left panel shows the location of the industry arrays (shaded polygons) and the SCSN broadband stations (black circles). The Long Beach array with 5300 stations (red dots) and the broadband station LGB (big red dot) that are used to create the correlations shown on the right. The correlations are done for approximately 500 hours, and are bandpassed  0.1-3.0 Hz.}
  \label{fig1} 
\end{figure*}

\section{Data: Real and Synthetic}

In this study, we use data from a temporary dense seismic network of 5340 stations in Long Beach, California \cite{lin2013high} that were originally used for an exploration survey conducted by an oil company. The broadband station LGB is part of the permanent earthquake monitoring array in the region (Southern California Seismic Network) and is cross-correlated with this array to form 5340 station pairs, from which we wish to determine the dispersion curves. The geometry of the array and a sample cross-correlation are shown in Figure 1.  The FTAN method was applied to each correlation pair for a range of frequencies between 0.2 and 5 Hz to construct images of dispersion curves. The band pass filter is a Gaussian filter $H(\omega)=\exp(-\alpha(\omega-\omega_0)^2/\omega_0^2)$ where we set filter parameter $\alpha=25$ for a compromise between the narrow-band assumption and filtering robustness, These were then hand-picked (labor-intensive) to create a set of labeled dispersion curves. We set aside 4340 of the curve images to be used as a testing set, and the rest (1000) were used as a training set for the method. In a production environment, we would hand-label only the 1000 training images and allow the algorithm to determine all remaining 4340 correlation pairs (testing set) without intervention. Here, we need to hand-label the testing set in order to assess our model’s final performance.

To limit the amount of labeled data necessary to train a model to pick dispersion curves, we designed an approach to generate realistic synthetic training data, with the ultimate goal of applying transfer learning to the trained model. To generate synthetic dispersion curves, we started with a 1D layered velocity model (Fig. 2) that matched the average dispersion curve for the region of the survey. This function was then perturbed both in velocities and layer thicknesses by a random amount up to 10\% of the original velocity function, forming an ensemble of different velocity models. The random variations on the layer thicknesses and velocities for this ensemble were drawn from a uniform distribution that centered at the starting 1D velocity model. For each instance, the dispersion curves were determined by a numerical solution of the eigen-problem \cite{aki2002quantitative,herrmann2013computer}. The resulting curves were then altered by random variations of up to +/-2.5\% in the frequencies and velocities and by adding random noise to the curves. In total, 100,000 synthetic curves were generated.

The synthetic and real dispersion curves are first pre-processed into image representations to make them suitable for convolutional networks. Initially, each curve is a collection of points with (frequency, velocity, amplitude) values such as shown in Figure 3. Each point also has an associated label from one of the three classes: fundamental mode, 1st overtone, and noise. We de-trend each curve in the frequency-velocity domain and transform the frequency axis to the logarithmic domain. Then, we create a 64-by-64 pixel greyscale image with pixel values between 0 and 255 representing the amplitude (Figure 3). To map individual points to the pixels, we treat each pixel as a discretized bin and fill its greyscale value as the amplitude of points that fall into the bin (post-detrending and log transform). The variable transforms are done to compact the information, thus reducing computational requirements. Similarly, the ground truth labels for each point of the dispersion curves are mapped to individual pixels by discretization and subsequent binning.

\begin{figure}
\centering
    \includegraphics[width=0.5\textwidth]{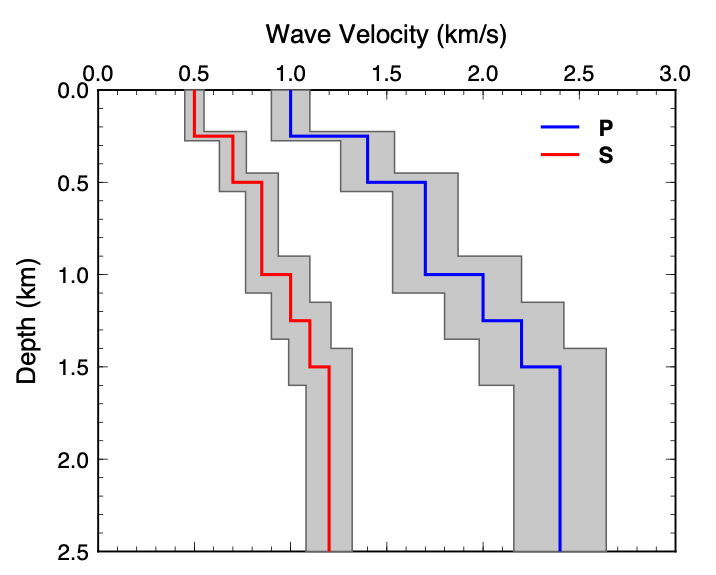}
\caption{Velocity Model. The P- and S-wave models used to construct the synthetic training set is shown along with the variations in velocity and layer depths.}
\label{fig2}
\end{figure}

\begin{figure*} 
    \centering
  \subfloat{%
       \includegraphics[width=0.45\linewidth]{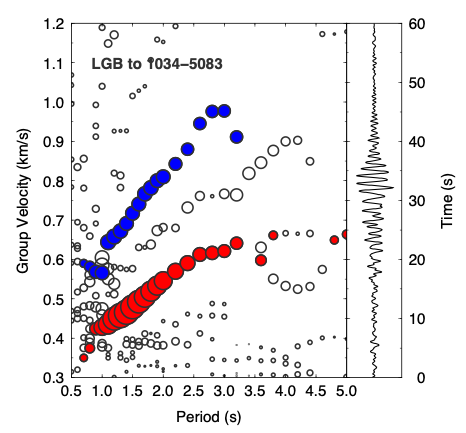}}
    \hskip -1ex
  \subfloat{%
        \includegraphics[width=0.43\linewidth]{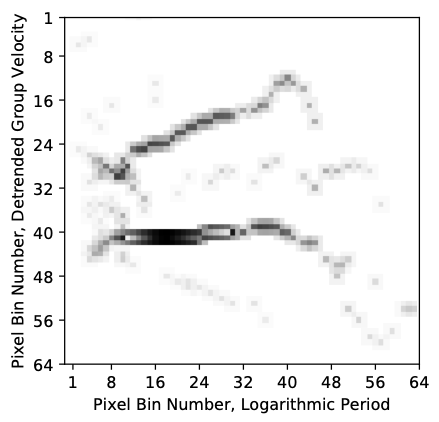}}
  \caption{Example of a hand-labeled dispersion curve for the path shown in Figure 1. The right panel shows the compacted image with logarithmic period axis and the linear trend removed. }
  \label{fig3} 
\end{figure*}

\section{Methods}
\subsection{Overview}
Our approach to picking dispersion curves uses deep convolutional networks in a supervised manner to perform pixelwise segmentation of the images. It consists of two main steps: (1) a U-Net architecture is trained first on the entirely synthetic dataset to learn coarse features, and (2) the best model is then fine-tuned to the limited amount of real data using a transfer learning approach. Below we describe each of these steps in detail.

The use of convolutional networks is well-motivated by the structure of our data, as the dispersion curve images exhibit spatially coherent geometric structure \cite{lecun2015deep}. Our problem is set up as one of fully-supervised image segmentation, since we have pixelwise labels for all images. The model used in this study is the U-Net architecture \cite{ronneberger2015u}, which is a deep convolutional network that has been successful for image segmentation tasks. In particular, the network applies a series of convolution and pooling layers to an input image to learn a sparse representation of it, and then applies a series of transpose convolution layers to finally output an image with the same lateral dimensions as the input. The depth of the output image is equal to the number of classes, which in our case is 3: fundamental mode, 1st-overtone, and noise.

Figure 4 provides a summary of the model used in this study. The network takes in a 64x64x1 image and outputs a stack of three images of equivalent dimension, with a softmax activation function applied to the outputs. In our case, the output of the neural network is a 64-by-64-by-3 array, with each pixel having 3 probabilities: [noise, fundamental mode curve, 1st overtone curve] associated with it. An example of an input image and the corresponding labels are shown in Figure 5 (upper panels).

To overcome the discretization error due to the 64x64 pixelization of the images, we post-process the picks by finding the corresponding closest frequency-time energy peaks in the original FTAN maps. The largely eliminates the errors introduced by the image formation process. Theoretically this could also be done by using finer grained images such as 128x128 pixels, but this would increase the computational requirements by a factor of 4.

\begin{figure*}
\centering
    \includegraphics[width=1.00\textwidth]{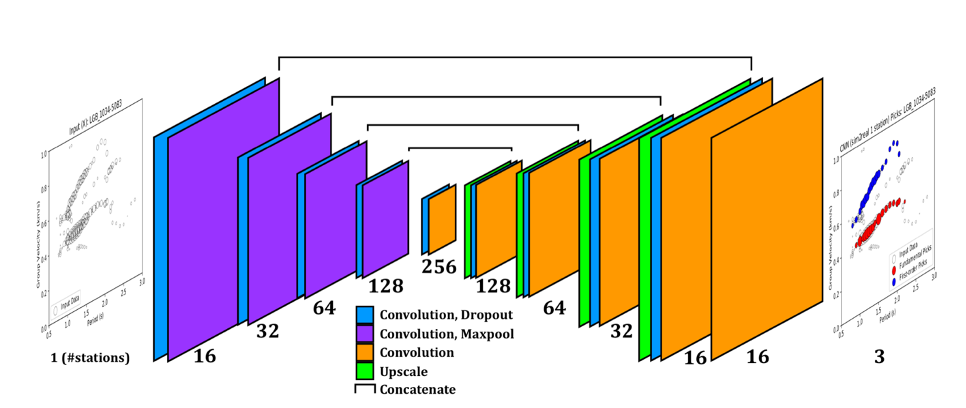}
\caption{A cartoon flowchart of how our data is processed throughout the paper. In the plots, the [X, Y, Size] axes represent [Period (s), Group Velocity (km/s), Amplitude], respectively. The center of the figure represents the convolutional neural network (CNN) structure that we employed. The goal of the paper is to take in a noisy plot and pick out points inside it that belong to the fundamental (red) and first-order (blue) overtones, while discarding everything else as noise.}
\label{fig4}
\end{figure*}

\subsection{Convolutional Neural Network Training}
Starting the training process using data from a simulation (a sim2real approach) avoids the need for extensive human labelling. We set aside 10\% of the synthetic images for model validation purposes and train the network on the remainder using the Adam optimizer \cite{kingma2014adam} using mini-batches of size 32. After each epoch of training, we check the model’s performance on the 10,000 unseen images in the validation set. If performance does not improve for 3 epochs in a row, we save the model after the best epoch.

Next, we proceed to fine-tune the best model on the synthetic data to the real Long Beach data using a transfer learning approach. Of the 5340 images from the Long Beach data set, we take a random subset of 1000 images for use in transfer learning. Of these 1000 images, 100 are used for validation (checking when to stop training the model) and 900 are used for updating model weights, which results in a 90\%-10\% train-validation split. For these 1000 images, we followed the same training procedure as with the synthetic data.
The remaining 4340 stations are reserved for evaluation of our models – we pretend that we have no access to their correct labels until after the final model is saved, as they are used only to determine whether our method is suitable for real usage.

Example output predictions are shown in Figure 5 along with the raw feature input and labels. It is clear that the model performs well for this examined image, correctly recovering nearly all fundamental and 1st overtone picks. Quantitative performance results on the validation set are provided in Figure 6, where precision and recall are computed for each of the three classes. The noise class has the highest precision and recall of the three classes (>99\%), which probably reflects the fact that the composition of the training dataset is heavily skewed toward noise. The fundamental and 1st-order modes have around 99\% and 98\% median precision, respectively, demonstrating that the model can accurately classify individual pixels. The median recalls for these classes are about 95\% and 94\%, respectively. The application of ML to this problem is able to substantially reduce the human-labor in analyzing surface wave data.

\begin{figure*} 
    \centering
  \subfloat{%
       \includegraphics[width=0.45\linewidth]{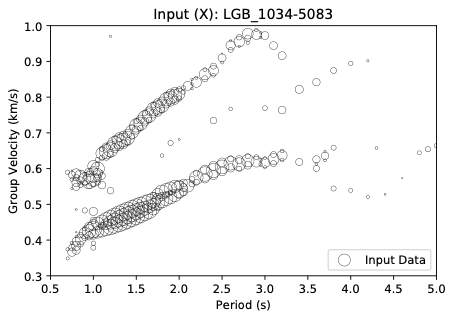}}
  \subfloat{%
        \includegraphics[width=0.45\linewidth]{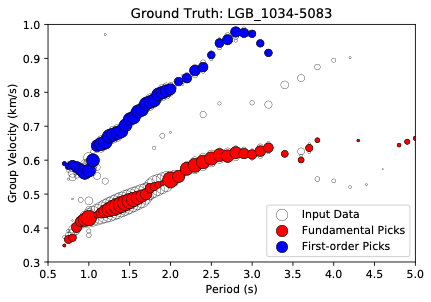}}
  \\
  \subfloat{%
        \includegraphics[width=0.45\linewidth]{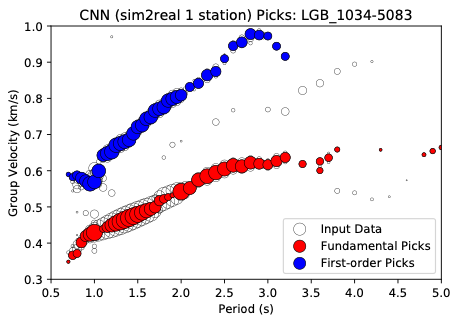}}
  \subfloat{%
        \includegraphics[width=0.45\linewidth]{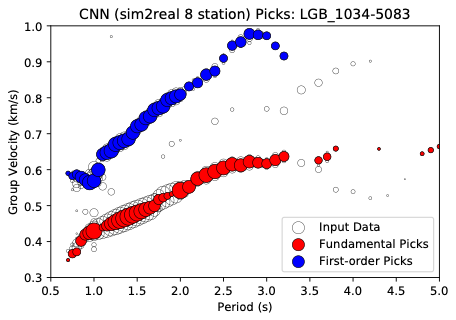}}
  \caption{Results for different algorithms. The top-left is the input data, which is converted to greyscale pixels. The top-right is a hand-picked classification of the dispersion curves. The bottom-left is the convolutional network prediction of labels using only one station input. The bottom-right is the results using 8 stations input.}
  \label{fig5} 
\end{figure*}

\subsection{Multi-station input}
The analysis described above was done for each station pair. We also explored the possibility of including neighboring stations into the feature set to better facilitate separation of genuine signal from noise. While the velocity structure may vary between different pairs, here we assume that these changes are small enough that the general characteristics from one dispersion curve to another are overall similar. The motivation is to use all of the available information together to make a decision, rather than examining one station pair at a time.

To do this, we include the images for $K$ nearest neighbor stations by concatenating them in the depth dimension to create a 3D input volume. Thus, the inputs are 64x64x$K$. We repeat the entire training procedure starting from generating synthetic data, as well as the transfer learning part. An example of a model using $K = 8$ is shown in Figure 5, which can be compared with the results for the $K = 1$ model seen previously. Figure 6 shows the median performance while increasing $K$ from 1 to 8. After accounting for the variability introduced by the stochastic nature of the training process (examining the 107 best training runs out of 215 total), we find that the performance does not improve significantly when adding in additional stations. We hoped that noise seen in one station might not be seen in a neighboring station, so a neural network might be able to combine multi-channel information to determine that this idiosyncratic noise is indeed noise. However, this unfortunately is not the case.

\begin{figure*}
    \centering
  \subfloat{%
       \includegraphics[width=0.45\linewidth]{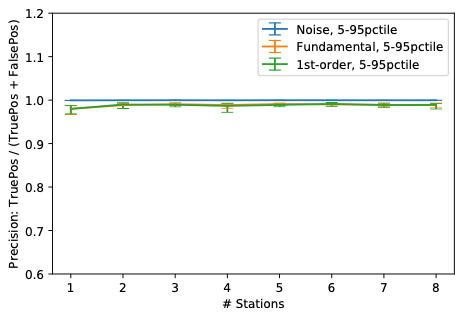}}
  \subfloat{%
        \includegraphics[width=0.45\linewidth]{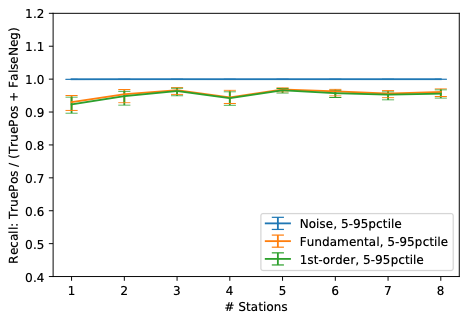}}
  \caption{Precision and Recall. Median precision and recall compared with the number of stations presented to the neural network simultaneously. $N=23$ (top 10\% of $N=239$ models in validation error) of synthetically trained, then Long-Beach trained models. Variations are due to changes in random seed for the model training, which changes the weight initializations and the order of training data shown. }
  \label{fig3} 
\end{figure*}

\section{Discussion}
The approach developed in this paper provides a means to train deep neural networks to perform dispersion curve picking using a hybrid simulation + real data scheme. The transfer learning step enables the neural network to learn coarse features from the simulation data that are also present in the real data, with the benefit that as much simulated data can be generated as needed. By then fine-tuning the model to the real data, the network learns finer scale features that are unique to these data, while only needing a relatively small amount of it. Here, we showed that just 1000 images are enough to adapt the initial model to the real data, although if more is available, the performance may improve further. This type of sim2real approach will likely be relevant to other problems in seismology where labeled data can be initially obtained from simulations.

We attempted many tweaks to the model hyperparameters as well as the randomization of workflow (same synthetic + model many real models, many synthetic models + many real models, varying learning rates, varying image dimensions, fixed vs. variable terminating epoch numbers, zero-padded empty channels vs. completely deleted empty channels, changing the probability threshold for picks, etc.) but failed to see any meaningful increase in performance as we increased station count. Therefore, it appears that there is no need to use more than 1 station for the proposed method, which also requires the least amount of training data.
We also tried to frame the problem as one of sequential classification, treating the picks as a sequence rather than converting to an image and using bidirectional Gated Recurrent Units (similar to Ross et al. \cite{ross2018p}). The sequences of floating-point tuples (period, group velocity, amplitude) directly extracted from the FTAN analysis were sorted by amplitude and presented to the neural network. This approach did not perform nearly as well as the convolutional network approach described above.

\begin{figure*}
\centering
    \includegraphics[width=0.85\textwidth]{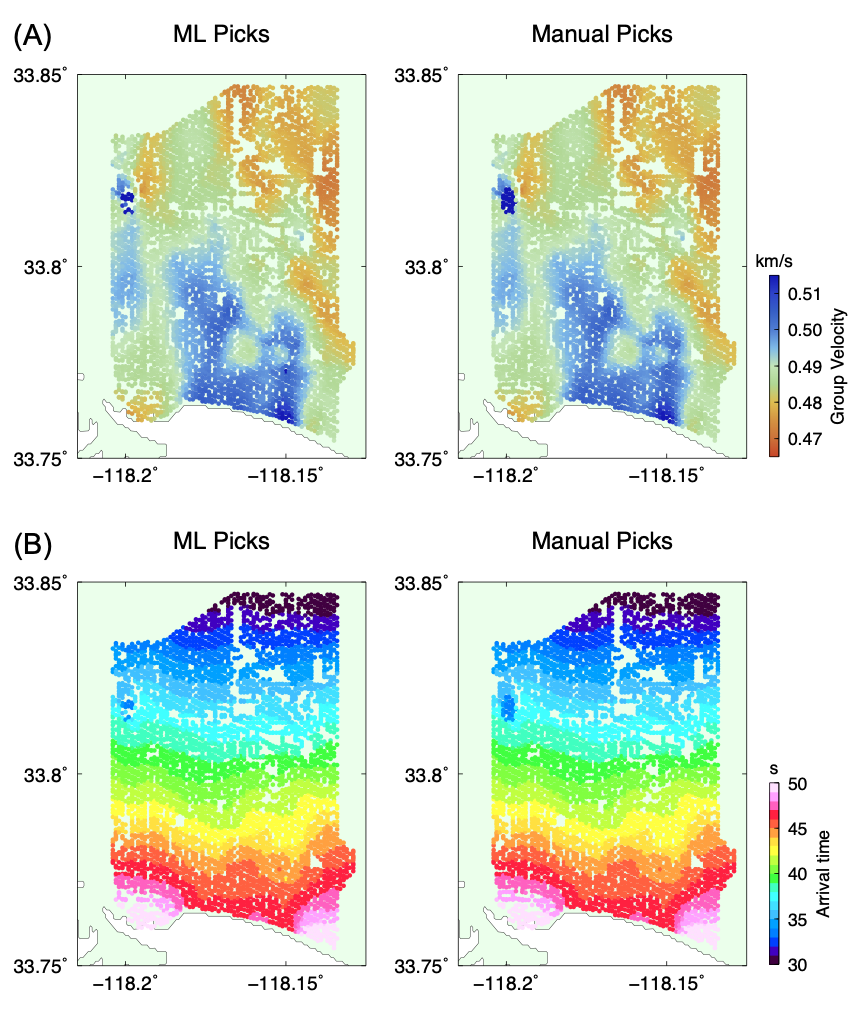}
\caption{Results for Long Beach Example. Shown are the group velocity maps (A) and the travel time picks (B) for the Long Beach data for the machine-learning processing (ML), and manual processing. That the two types of processing produce nearly identical results indicates that the machine-learning approach is working.}
\label{fig2}
\end{figure*}

\section{Conclusions}
We have developed a machine learning method for extracting dispersions curves from velocity-frequency images. The procedure was training with a combination of synthetic examples and labelled real data. Testing on real data shows the method works with a median per-class precision of at least 98\% and a per-class recall rate of at least 94\%. We achieved an 80\% reduction in human labor using this extraction technique on a dataset of 5340 curve sets, and we expect this efficiency to improve further if applications on future datasets start with this pretrained model. The value of this method is its ability to be applied in bulk, and will be more apparent as more datasets are used.

With this new method to classify points for dispersion curve fitting, it is now possible to ingest large volumes of recorded sensor data with minimal human input, and then systematically calculate travel times as shown in Figure 7. Previously, this data ingestion step for each dataset would take hours of human work consisting of point selection by heuristics and experience, but it can now be taken care of with a pipeline designed to separate noise from dispersion curve.

\section*{Acknowledgment}
We thank Signal Hill Petroleum for permission to use the Long Beach Array, and the Southern California Seismic Network for providing data from the broadband stations. We also thank Yisong Yue for helpful discussions. This study was partially supported by NSF/EAR 1520081.

\ifCLASSOPTIONcaptionsoff
  \newpage
\fi



%
\bibliography{refs}
\bibliographystyle{IEEEtran}

\clearpage

%

\begin{IEEEbiography}[{\includegraphics[width=1in,height=1.25in,clip,keepaspectratio]{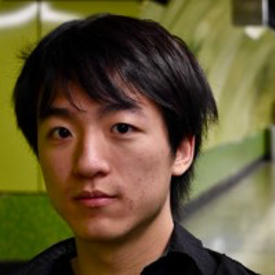}}]{Xiaotian (Jim) Zhang}
is a student of the Computing and Mathematical Sciences Department who works in the areas of machine learning applications and probability models. He has created models and performed research at Caltech as well as various industry firms.
\end{IEEEbiography}
\vskip -1ex
\begin{IEEEbiography}[{\includegraphics[width=1in,height=1.25in,clip,keepaspectratio]{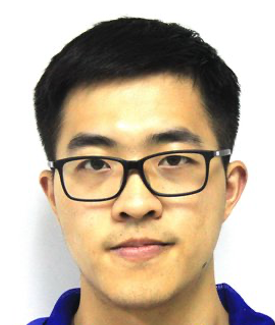}}]{Zhe Jia}
is a graduate student in geophysics at the Seismological Laboratory, Caltech. He major interests lie in characterizing rupture processes of complex earthquakes and studying the shallow structure with ambient noise tomography.
\end{IEEEbiography}
\vskip -1ex
\begin{IEEEbiography}[{\includegraphics[width=1in,height=1.25in,clip,keepaspectratio]{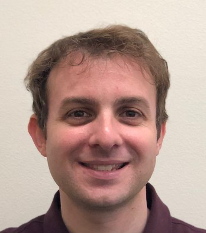}}]{Zachary E. Ross}
is an Assistant Professor of Geophysics who uses machine learning and signal processing techniques to better understand earthquakes and fault zones. He is interested in seismicity, earthquake source properties, and fault zone imaging.
\end{IEEEbiography}
\vskip -1ex
\begin{IEEEbiography}[{\includegraphics[width=1in,height=1.25in,clip,keepaspectratio]{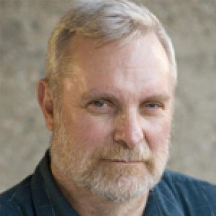}}]{Robert W. Clayton}
is a Prof. of Geophysics who works in the areas of seismic wave propagation, earth structure and tectonics. He has applied imaging methods to the Los Angeles region and to subduction zones around the world.
\end{IEEEbiography}



\end{document}